\documentclass[letterpaper, 10pt, conference]{ieeeconf} 
\IEEEoverridecommandlockouts
\usepackage{cite}
\usepackage{amsmath,amssymb,amsfonts}
\usepackage{algorithmic}
\usepackage{graphicx}
\usepackage{textcomp}
\usepackage{xcolor}
\usepackage{todonotes}

\usepackage{todonotes}
\usepackage{soul}
\definecolor{smoothgreen}{rgb}{0.7,1,0.7}
\sethlcolor{smoothgreen}

\def\BibTeX{{\rm B\kern-.05em{\sc i\kern-.025em b}\kern-.08em
    T\kern-.1667em\lower.7ex\hbox{E}\kern-.125emX}}

\title{\Huge \bf Social Navigation for Tour-guide Robot}

\author{Vikram Shree$^{*,1,2}$ and Jose Nino$^{*,1,3}$
\thanks{$^{*}$Equal contribution.}
\thanks{$^{1}$Work done at the department of Mechanical and Aerospace Engineering, Cornell University, Ithaca NY (USA).  Email: {\tt \{vs476, jan268\}@cornell.edu}, respectively.}
\thanks{$^{2}$SiLC Technologies, Monrovia CA (USA).}
\thanks{$^{3}$Draper, Cambridge MA (USA).}
\thanks{The manuscript has been accepted at the 5th Workshop on Social Robot Navigation, held at the IEEE/RSJ International Conference on Intelligent Robots and Systems (IROS) 2026.}
}

\begin{document}

\maketitle


\begin{abstract}
We propose a force-based model for social navigation of a tour-guide robot. Social forces due to various factors like obstacles, user position and heading, have been accounted for in the model. We claim that each one of these forces makes the robot more sociable to the user and we design an experimental setup for evaluation. In the experiment, the user follows an autonomous robot to a destination in a known map, while undertaking a few simple sub-tasks in the middle, which serve as distractions. For each participant, we run several rounds of the experiment, each with different forces and a shortest path, A*-search baseline model. Using per-round subjective indicators, we propose to study the effect of our force model on constructs such as: follow-ability, perceived safety, and perceived intelligence.

\end{abstract}


\section{Introduction}
\label{sec:introduction}
In this work, we present a social force model for a tour-guide robot and conduct experiments to study its effect on human’s perception about the robot. As the globe is becoming more connected, cities around the world are attracting more tourists every year. As a result, more people are exploring public spaces like museums, parks and shopping malls. Often, these huge public spaces are clumsy and time consuming to navigate through, especially for elderly people. This motivates the development of a tour-guide robot, which helps the tourists in navigating to their destinations, thus, making the tour more fulfilling and productive. 
 
To be a successful tour-guide robot, it needs to properly follow the expectations of tourists, thus adding social constraints on robot’s behavior. For example, it is reasonable for a tourist to expect that the robot moves at a safe speed, so that they can follow it without rushing through the scene. Thus, traditional shortest path planning approaches need to be adapted in this scenario to account for social constraints imposed by the user. In this work, we are interested in evaluating the benefits of following social constraints from a human's perspective.
 
\begin{figure}
    \centering
    \includegraphics[trim= 0 0 0 0, clip, width=.40\textwidth]{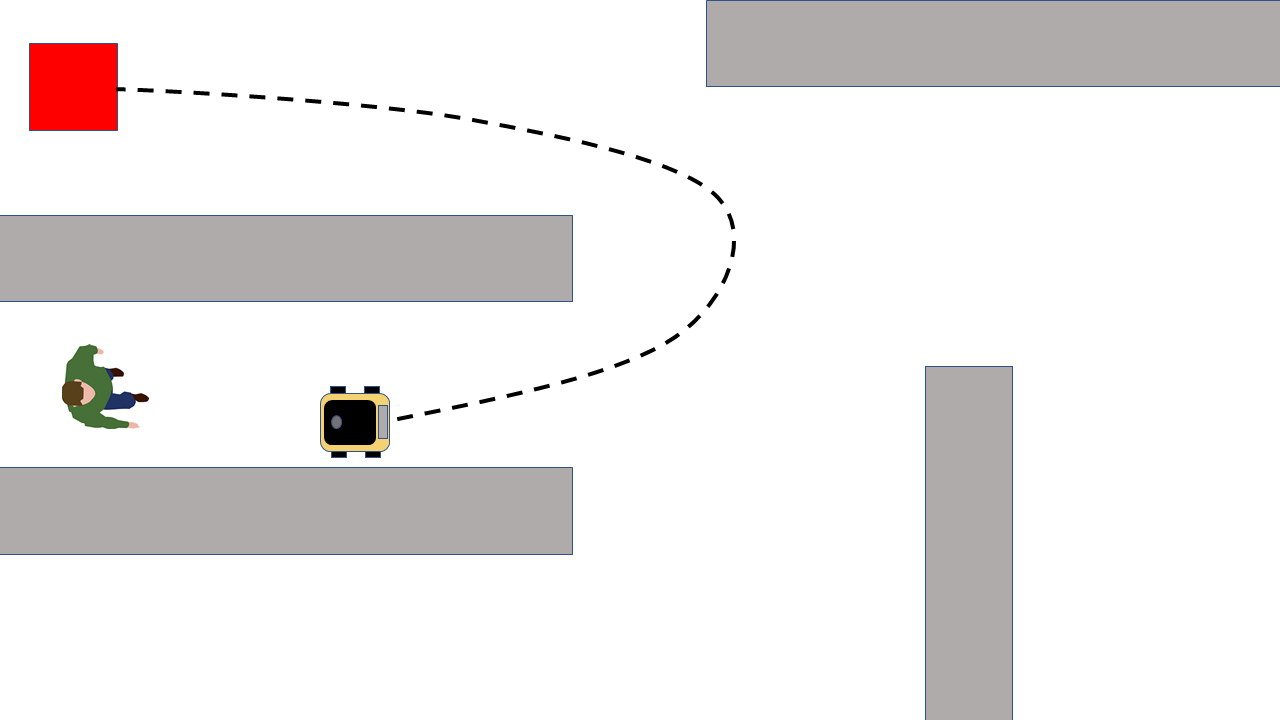}
    \caption{\small Schematic of a tour-guide robot helping user to navigate to goal, shown in red.}
    \label{fig:overall_task}
\end{figure}

The main focus of social navigation research in the past years has been to help the robot stay farther away from humans in crowded environments; however, there is little work in the case when the robot has to guide a human to a destination. In scenarios, where the robot has to simply reach a destination, the humans are treated as obstacles and thus, there is a penalty associated with coming too close to the humans. In contrast, a tour-guide robot needs to stay close to the tourist, so that they do not get lost in the crowd. Furthermore, it should move in a manner that makes it easier for the tourist to follow. These challenges motivate our social force model where the robot is acting as the leader and is tasked to guide the tourist to their destination, in a friendly manner. Fig.~\ref{fig:overall_task} shows schematic of a social tour-guide robot, helping out a user to navigate safely to the goal location, while following social constraints.

\section{Related Work}
\label{sec:related_work}

\subsection{Robot Navigation in Presence of Humans}
Social constraints first entered into navigation with a promise of keeping safe distance between the robot and humans \cite{bellarbi2016social, ramon2014transferring, perez2014robot}. Bellarbi et al. \cite{bellarbi2016social} established social rules that allow the robot to navigate smoothly and socially in indoor environments. Ramon-Vigo et al. \cite{ramon2014transferring} leveraged inverse reinforcement learning to transfer skills of human navigation behavior to a robot. Perez-Higueras et al. \cite{perez2014robot} take a cost-based optimization route where a social cost map is learned from expert demonstrations. However, these works do not address the challenges that arise when the robot is leading a human, which is necessary for a tour-guide robot. 

Few researchers have studied the problem of human interaction with guide robots \cite{shen2006visual, azenkot2016enabling}. Shen et al. \cite{shen2006visual} explored the ``object approaching'' behavior for guide robots in a museum. Azenkot et al. \cite{azenkot2016enabling} introduced multiple modes of guidance for blind people. However, these works focused on human-robot communication, and social navigation constraints were ignored.

\subsection{Social Force Model}
Researchers first proposed to add pseudo forces onto the robot to mimic socially acceptable behavior \cite{helbing1995social, zhang2016framework}. One of the earliest works on social forces is by Helbing et al. \cite{helbing1995social} where the authors added social forces owing to the presence of humans and intermediate goals in the environment. Zhang et al. \cite{zhang2016framework} incorporate attractive and repulsive forces for the robot to address two social constraints: speed control to stay close to the user and maintaining safe distance from obstacles in the environment. However, they fail to evaluate the effect of those forces through human experiments and surveys.

Our work extends the existing social force model for tour-guide robot, developed by Zhang et al. \cite{zhang2016framework}, by adding additional force due to the limited field of view of the human. 
In addition to adapting the speed and maintaining safe distance from obstacles, we add attractive force in order to impel the robot to stay close to the heading direction of the user. Furthermore, to establish the utility of the force model, we evaluate the effect of each force on the user through rigorous human-participant studies.


\section{System}
\label{sec:system}

To behave socially, we believe that a tour-guide robot should adapt to the motion of the user, so that it can suit their desires and needs. We leverage potential fields for achieving this adaptive control for the robot. In this work, we propose to generate adaptive potential fields due to static and dynamic obstacles, the tourist and intermediate goal points. The robot tends to move in the direction of lower potential, thus producing the required adaptive behavior.

\subsection{Potential Field}
\label{subsec:pot_field}
We apply two types of potential fields: attractive and repulsive. These fields get updated online based upon the configuration of the world and forces are generated accordingly to control the robot.

\subsubsection{Sub-goal Attractive Potential}
\begin{figure}[h]
    \centering
    \includegraphics[trim= 75 40 65 50, clip, width=.40\textwidth]{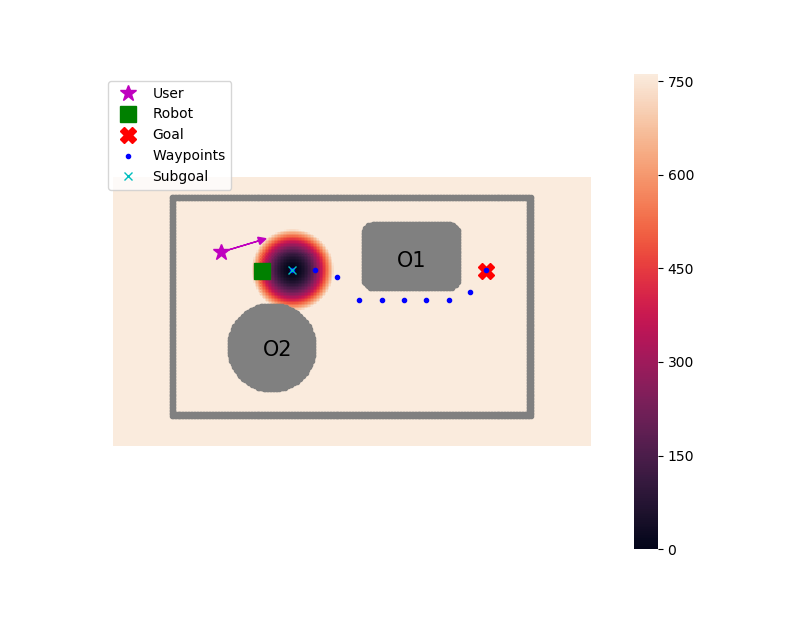}
    \caption{\small Potential field generated from closest waypoint to the robot. }
    \label{fig:goal_attract}
\end{figure}

To force the robot to move towards a nearby goal point, similar to previous work by Zhang et al. \cite{zhang2016framework}, we apply an attractive potential from it. Often this goal is not the final destination for the robot, rather an intermediate subgoal on the path to the final destination. This is to avoid the robot getting stuck in dead-lock situations. The attractive potential $P_{ag}$ is given by:
\begin{equation}
    P_{ag} = \frac{K_{ag}}{2} \times d_{g}^{2}
    \label{eq:goal_attract_pot}
\end{equation}
Where $K_{ag}$ is a constant coefficient and $d_g$ is the distance of the robot from the goal. Fig.~\ref{fig:goal_attract} shows the sub-goal and the corresponding attractive potential field for a model environment. This causes an attractive force $F_{ag}$ to act on the robot which is given by: 
\begin{equation}
    F_{ag} = - K_{ag} \times d_{g}
    \label{eq:goal_attract_force}
\end{equation}
The proportional force allows the robot to drive faster towards the goal, when it is far away from the goal.

\subsubsection{Obstacle Repulsive Potential}
\begin{figure}[h]
    \centering
    \includegraphics[trim= 75 40 65 50, clip, width=.40\textwidth]{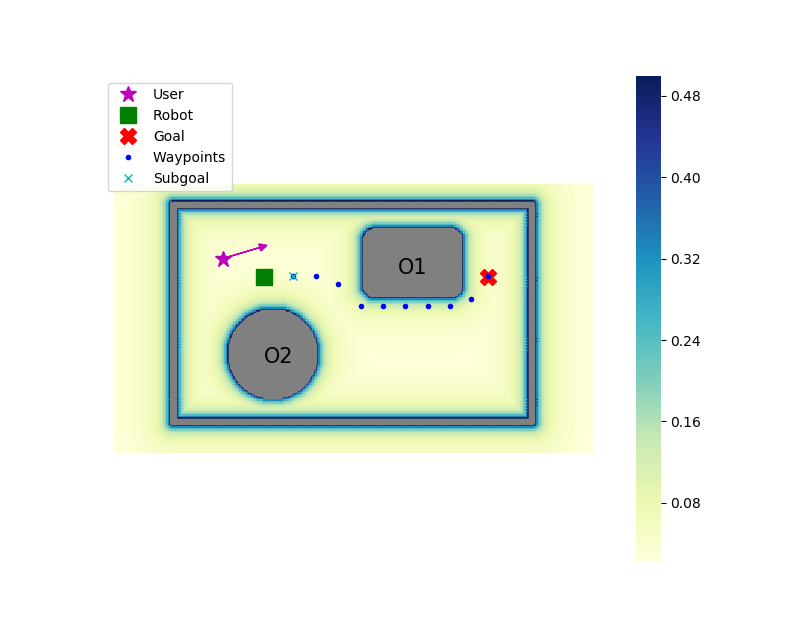}
    \caption{\small Potential field generated from the obstacles.}
    \label{fig:obstacle_repel}
\end{figure}

To avoid collision with static and dynamic obstacles in the map, repulsive potential is developed around them in the same way as in prior work by Zhang et al. \cite{zhang2016framework}. The repulsive potential $P_{ro}$ is given by:
\begin{equation}
    P_{ro} = \frac{K_{ro}}{(d_{r}-d_{o})} \ \text{if} \ d_r > d_o; \ \ P_{ro} = \infty, \ \text{otherwise}.
    \label{eq:obstacle_repel_pot}
\end{equation}
$K_{ro}$ is a constant coefficient, $d_r$ is the distance of the closest point on the obstacle to the robot and $d_o$ is a parameter defining the minimum distance that the robot can approach the obstacle, without hitting it. Fig.~\ref{fig:obstacle_repel} shows the obstacles ($O1, O2$) in a model environment and their corresponding repulsive potential fields. This causes a force $F_{ro}$ to act on the robot which is given by:
\begin{equation}
    F_{ro} = \frac{K_{ro}}{(d_{r}-d_{o})^{2}} \ \text{if} \ d_r > d_o; \ \ F_{ro} = \infty, \ \text{otherwise}.
    \label{eq:obstacle_repel_force}
\end{equation}
In general, the path of the robot does not intersect with any obstacle. However, having the repulsive force puts an additional guarantee to drive the robot away, in case it comes closer to the obstacles. In addition, the inverse quadratic relationship ensures that the robot’s path is unaffected by the repulsive force, if it is sufficiently far away from any obstacles.

\subsubsection{Attractive-Repulsive Potential due to User}
\begin{figure}
    \centering
    \includegraphics[trim= 75 40 65 50, clip, width=.40\textwidth]{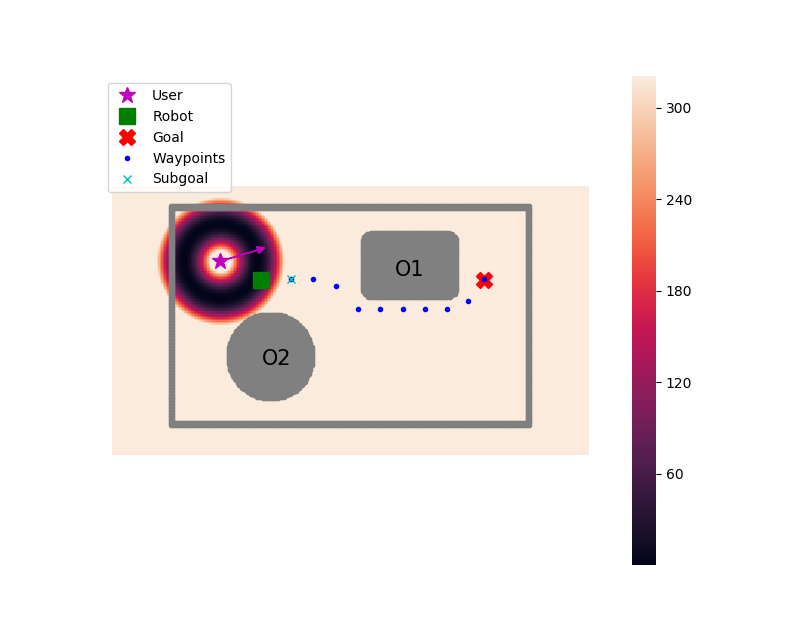}
    \caption{\small Potential field generated from the User.}
    \label{fig:user_attract_repel}
\end{figure}

For adapting the robot to the motion of the user, similar to the prior work by Zhang et al. \cite{zhang2016framework}, we apply a special potential field that has both, attractive and repulsive, properties in it. The potential field $P_{ar}$ is given by:
\begin{equation}
    P_{ar} = \frac{K_{ar}}{2} \times (d_s - d_u)^2 
    \label{eq:obstacle_repel_pot}
\end{equation}
Where $K_{ar}$ is a constant coefficient, $d_s$ is a parameter that can be viewed as the social distance and $d_u$ is the distance of the robot from the user. Fig.~\ref{fig:user_attract_repel} shows a user in a model environment and their corresponding potential field $P_{ar}$. This leads to a force $F_{ar}$ on the robot which is given by:
\begin{equation}
    F_{ar} = K_{ar} \times (d_s - d_u) 
    \label{eq:obstacle_repel_pot}
\end{equation}
If the robot gets too far ahead of the user, the attractive force slows it down. The repulsive force gets activated, when the robot comes too close to the user, thus driving it away from the user. As a result, this combination of attractive-repulsive potential fields from the user helps the robot to maintain a desirable relative distance from the user.

\subsubsection{Attractive Potential due to User's Field of View}
\begin{figure}[h]
    \centering
    \includegraphics[trim= 40 220 430 15, clip, width=.22\textwidth]{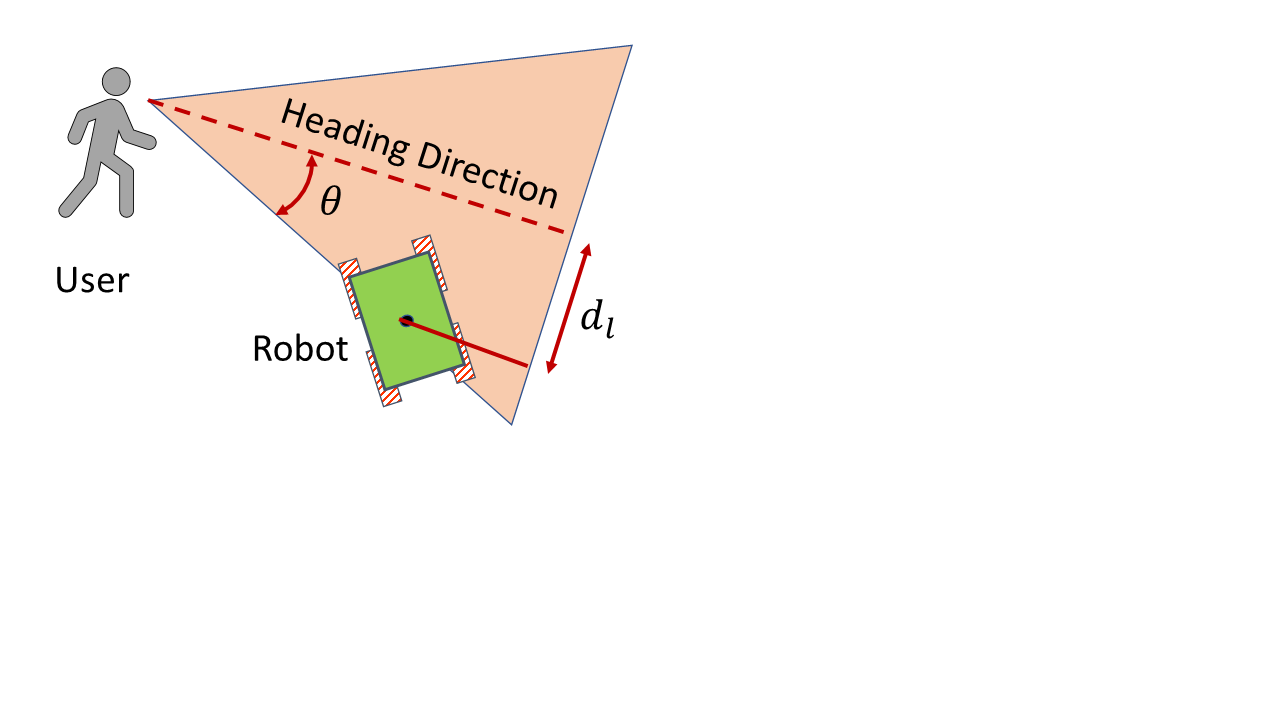}
    \caption{\small Field of view of a user.}
    \label{fig:fov_schematic}
\end{figure}

\begin{figure}
    \centering
    \includegraphics[trim= 75 40 65 50, clip, width=.40\textwidth]{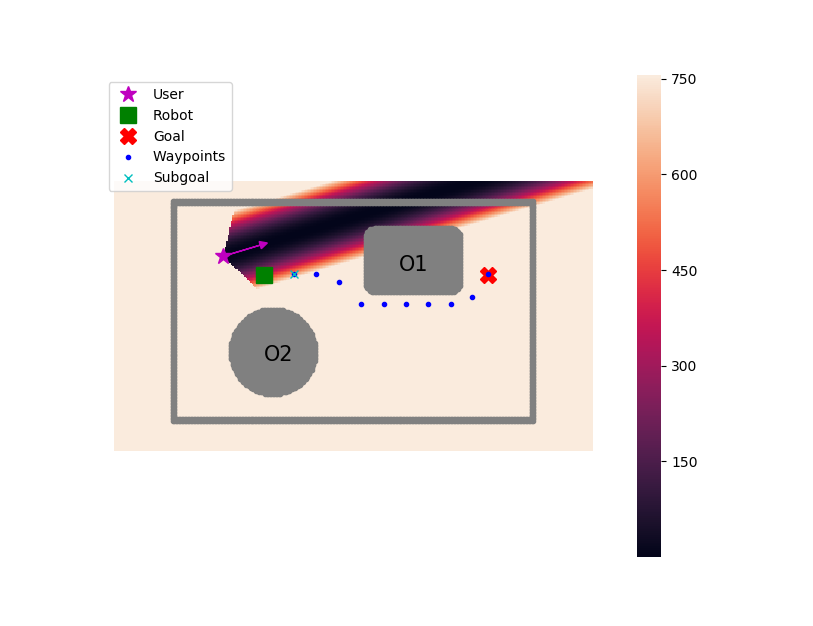}
    \caption{\small Potential field generated from user's field of view.}
    \label{fig:fov_attract}
\end{figure}

To keep the robot in the field of view of the user, we add an attractive potential that pulls the robot towards the heading direction of the user, as depicted in Fig.~\ref{fig:fov_schematic}. The potential field $P_{fov}$ is given by:
\begin{equation}
    P_{fov} = \frac{K_{fov}}{2} \times d_{l}^{2}, \ \text{if} \ \theta < \theta_{o}.
    \label{eq:fov_attract_pot}
\end{equation}
Where $K_{fov}$ is a constant coefficient, $d_l$ is the distance of the robot from the user's heading direction, $\theta$ is orientation of the vector from user to robot, and $\theta_{o}$ is half of the maximum field of view. Fig.~\ref{fig:fov_attract} depicts the field of view of a user and the corresponding potential associated with it for a model environment. The force $F_{fov}$ acting on the robot is:
\begin{equation}
    F_{fov} = -K_{fov} \times d_{l}, \ \text{if} \ \theta < \theta_{o}; \ F_{fov} = F_{\text{max}}, \ \text{otherwise}.
    \label{eq:fov_attract_force}
\end{equation}
The force due to the field of view constraint makes it easier for the user to keep track of the robot by driving the robot towards the heading direction of the user.

\subsection{Global Planner}
\label{subsec:glob_planner}
Apart from the local forces experienced by the robot, we utilize A* algorithm \cite{Hart1968} to generate the global path from robot’s current location to the goal location. Typically, it is desirable to guide the user to the destination in a minimum amount of time. Accordingly, the A* algorithm produces a series of contiguous points corresponding to the shortest path in the discrete map. These points are treated as subgoals for the robot, eventually leading to the final destination.

\subsection{Implementation}
\label{subsec:implementation}
We design our system in a manner so that we can conduct our study in-person or virtually. There are slight differences in the experimental procedure depending on which method is used.

\subsubsection{Simulation Environment}

\begin{figure}
    \centering
    \includegraphics[width=.30\textwidth]{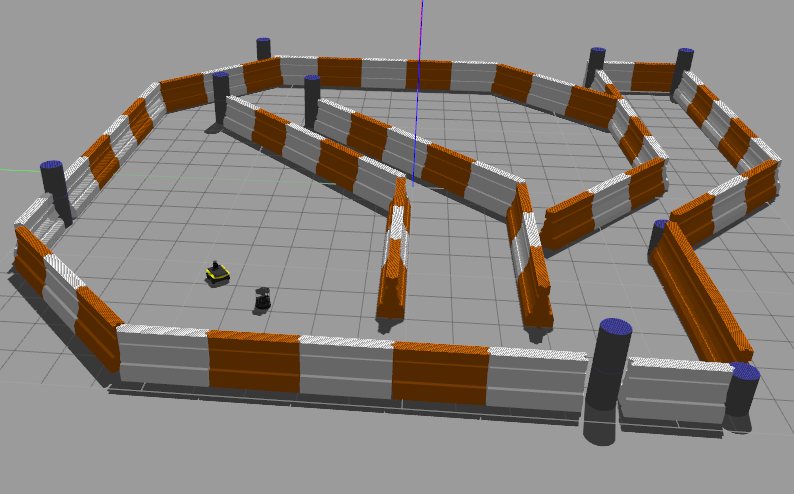}
    \caption{\small Gazebo environment for conducting virtual experiments, with Jackal robot and Turtlebot3 in it.}
    \label{fig:gazebo_world}
\end{figure}

\begin{figure}
    \centering
    \includegraphics[width=.21\textwidth]{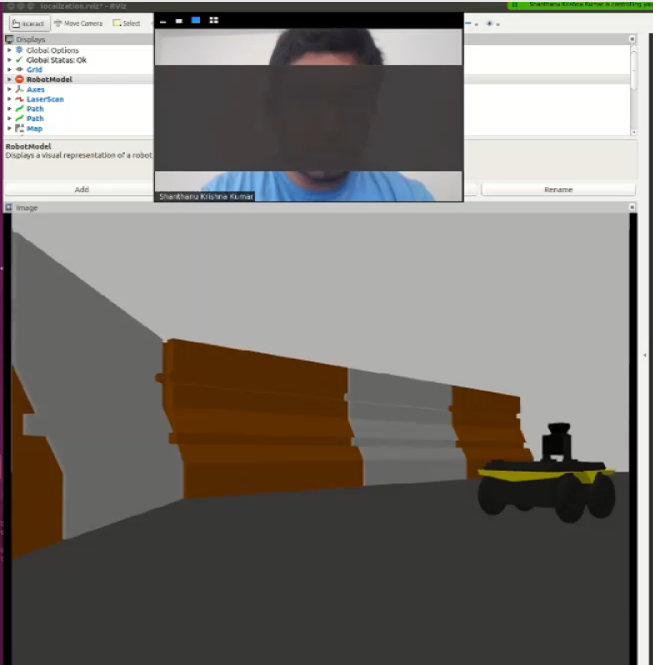}
    \caption{\small Zoom study setup. Participant is given access to the computer running ROS, through zoom. Using the keyboard and images from Turtlebot3, they follow the Jackal robot to the goal location. }
    \label{fig:study}
\end{figure}

For the virtual study, we implemented our system in Gazebo and ROS kinetic. We use a ClearPath Jackal robot as the leader robot. The Jackal is equipped with RGBD sensors and a 16-scan lidar. We modified ROS packages from ClearPath to create a Gazebo world map and leverage the basic navigation stack. These packages were also modified to include a Turtlebot3 robot which is used as an avatar for the human follower. Fig.~\ref{fig:gazebo_world} shows the simulation environment with the robots in it. The open source Turtlebot3 navigation stack is used. The Turtlebot3 is equipped with an RGB camera and all other sensors are deactivated. Global position and orientation of both the Jackal and Turtlebot3 are tracked using the Gazebo's state messages. Additionally, during the virtual study we use zoom to give participants control of our keyboard and access to our display. Participants control the Turtlebot3 through a keyboard using the open source {\tt teleop\_twist\_keyboard} package. They are shown the RGB video feed from the front camera of Turtlebot3 using {\tt rviz} as seen in Fig.~\ref{fig:study}. Participants are not allowed to see the overhead Gazebo map during the study. The social forces and A* algorithms are both implemented in Python language.

\subsubsection{Real-world Implementation}
For the in-person study, the implementation is similar. However, we remove the Turtlebot3 as the person can follow the robot themselves. One major change is the need for a lidar detector for pedestrians and a Kalman filter tracker to infer the participant’s position and orientation. The position and orientation of the jackal is tracked using ClearPath’s own navigation stack. The Jackal’s navigation stack includes a map making package which is used to globally locate stationary obstacles in the study environment. All other aspects of the implementation are shared between the two studies.

\section{Evaluation Method}
\label{sec:evaluation_method}

\subsection{Hypotheses}
The purpose of this study is to measure the influence of each social force on the following constructs: follow-ability, perceived safety, and perceived intelligence when compared to an A*-search baseline model. Therefore, for each social force ``x" we propose the following hypotheses:

\textbf{H1}: A robot that uses social force x is perceived as more intelligent than the baseline.

\textbf{H2}: A robot that uses social force x is perceived as safer than the baseline.

\textbf{H3}: A robot that uses social force x is perceived as easier to follow than the baseline.

\subsection{Procedure}
For the study we ask participants to follow the Jackal robot to four goal positions in the map. The robot navigates to each goal point while using one of the following algorithms: baseline, obstacle force ($F_{ro}$), Field of view force ($F_{fov}$), and proximity force ($F_{ar}$). We refer navigation to one goal as a \textit{following} sequence. For each \textit{following} sequence, participants and the robots are placed at a predefined start position. The robot is then commanded to navigate to one of the pre-selected goal points. The goal point and social force algorithm used during the following sequence is randomly selected to avoid ordering effects. During a \textit{following} sequence, the participant is given a subtask of completing math problems to simulate distractions in the environment. At the end of the following sequence the participants and robot are reset to the start position.

To measure the impact of the social forces on the constructs of follow-ability, perceived safety, and perceived intelligence the participants are asked to complete a customized Godspeed Questionnaire \cite{godspeed}. The Godspeed questionnaire is a well-used metric in Human robot interaction studies and examines if participants associate the following factors with the robot: anthropomorphism, animacy, likability, perceived intelligence and perceived safety. Instead of using all factors within the questionnaire we chose to only use characteristics under perceived intelligence and perceived safety. Additionally, we choose humanlike and elegant motion characteristics from anthropomorphism to represent the construct of follow-ability of the robot. The survey questions and associated constructs can be seen in Table~\ref{tabel:survey}. This survey is based on a 1 to 5 Likert scale. We compare the results for each social force against the baseline. We use one-tailed paired t-test to test the null hypothesis for \textbf{H1}, \textbf{H2}, and \textbf{H3}. If the resultant p-value for a hypothesis is less than 0.05 then we can reject the null hypothesis and accept the alternative hypothesis.
We also measure the Cronbach's alpha for each construct to ensure internal consistency. If the Cronbach’s alpha is above 0.7 then we can say the survey is internally consistent.

\begin{table}
\centering
\caption{\small Modified Godspeed survey questions and associated constructs.}
\begin{tabular}{|l|l|} 
\hline
Survey Questions (1-5 scale)    & Construct               \\ 
\hline
Machinelike/Humanlike           & Follow-ability~          \\ 
\hline
Fake/natural                    & Follow-ability~          \\ 
\hline
Moving rigidly/moving elegantly & Follow-ability~          \\ 
\hline
Incompetent/competent           & Perceived Intelligence  \\ 
\hline
Ignorant/Knowledgeable          & Perceived Intelligence  \\ 
\hline
Irresponsible/Responsible       & Perceived Intelligence  \\ 
\hline
Unintelligent/Intelligent~      & Perceived Intelligence  \\ 
\hline
Foolish/Sensible                & Perceived Intelligence  \\ 
\hline
Anxious/Relaxed                 & Perceived Safety        \\ 
\hline
Agitated/Calm                   & Perceived Safety        \\ 
\hline
Quiescent/Surprised             & Perceived Safety        \\
\hline
\end{tabular}
\label{tabel:survey}
\end{table}

\subsection{Participants}

We plan to recruit participants from Cornell University for the study. Each participant, after completion of the experiment receives a \$10 Amazon gift coupon. To determine the number of participants, we conducted a power analysis test. For this analysis, we need to set the power, significance level, and effect size. 
To avoid a type-2 error most literature picks a value of 0.8 or above for the power. In our case we chose a value of 0.95. 
We choose the commonly used significance level of $0.05$.
The effect size is a measure of the difference between two groups. This can usually be found through a pilot study or through literature review of similar studies. 
Based on Cohen's recommendation \cite{Cohen1988}, we have chosen a medium effect size of 0.5, which yields a participant count of 44 for our study.



\section{Discussion}
\label{sec:discussion}


Through our simulations, we observed that the behavior of the Jackal can vary greatly based on the selected parameters for each force. For example, the jackal can get stuck in a local minimum if the goal and obstacle forces cancel each other. Additionally, if the repulsive force from the person is too strong then the robot can miss goal locations or crash into an obstacle. For this reason, these parameters had to be carefully tuned to avoid such pathological behavior. An extension of this work could be to learn these parameters through reinforcement learning through observations of human controlled robots. 

Our virtual setup provides a convenient opportunity to conduct experimental studies in situations where the participants cannot come to laboratory for controlled experiments. However, there are a few caveats. The pilot study elucidated a few of these challenges. The user felt difficulty in controlling their Turtlebot3 avatars because of the discrete controller behavior. Furthermore, the small field of view of the Turtlebot3 makes the user feel more restricted, thus affecting the results of our survey. Therefore, to perform a more sincere evaluation of the force-model, we believe that the in-person study is more suitable.

\section{Conclusion}
\label{sec:conclusion}

In this work we implemented a social force-based model which accounts for obstacles, social spacing, and field of view of the human follower. We designed a virtual human-participant study setup to measure the influence of social forces on follow-ability, perceived safety, and perceived intelligence. Based on a few trials, we received constructive feedback in our virtual, Zoom study; however, the reported difficulty of navigating the Turtlebot3 robot may influence our results and therefore future work will focus more on the in-person study to avoid such issues.

\section*{Acknowledgment}

We thank Prof. Guy Hoffman for his valuable feedback throughout the project and Shanthanu Krishnakumar for participating in the pilot test.

\vspace{12pt}
\end{document}